\documentclass{article}

\usepackage{microtype}
\usepackage{graphicx}
\usepackage{subfigure}
\usepackage{booktabs} % for professional tables
\usepackage{amsmath}
\usepackage{amssymb}

\newcommand{\R}{\mathbb{R}}
\usepackage{hyperref}

\usepackage[accepted]{icml2019}

\icmltitlerunning{Unsupervised Representations of Pollen in Bright-Field Microscopy}

\begin{document}

\twocolumn[
\icmltitle{Unsupervised Representations of Pollen in Bright-Field Microscopy}

% It is OKAY to include author information, even for blind
% submissions: the style file will automatically remove it for you
% unless you've provided the [accepted] option to the icml2019
% package.

% List of affiliations: The first argument should be a (short)
% identifier you will use later to specify author affiliations
% Academic affiliations should list Department, University, City, Region, Country
% Industry affiliations should list Company, City, Region, Country

% You can specify symbols, otherwise they are numbered in order.
% Ideally, you should not use this facility. Affiliations will be numbered
% in order of appearance and this is the preferred way.
\icmlsetsymbol{equal}{*}

\begin{icmlauthorlist}
\icmlauthor{Peter He}{equal,icl}
\icmlauthor{Gerard Glowacki}{hch}
\icmlauthor{Alexis Gkantiragas}{equal,ucl}
\end{icmlauthorlist}

\icmlaffiliation{icl}{Department of Computing, Imperial College London}
\icmlaffiliation{ucl}{Department of Molecular Biology, University College London}
\icmlaffiliation{hch}{Hampton Court House, London, United Kingdom}

\icmlcorrespondingauthor{Peter He}{peter.he18@imperial.ac.uk}
\icmlcorrespondingauthor{Alexis Gkantiragas}{alexis.gkantiragas.17@ucl.ac.uk}

% You may provide any keywords that you
% find helpful for describing your paper; these are used to populate
% the "keywords" metadata in the PDF but will not be shown in the document
\icmlkeywords{Machine Learning, ICML, Pollen Analysis, Manifold Learning, Unsupervised Learning, Computational Biology, Bright-Field Microscopy}

\vskip 0.3in
]

% this must go after the closing bracket ] following \twocolumn[ ...

% This command actually creates the footnote in the first column
% listing the affiliations and the copyright notice.
% The command takes one argument, which is text to display at the start of the footnote.
% The \icmlEqualContribution command is standard text for equal contribution.
% Remove it (just {}) if you do not need this facility.

%\printAffiliationsAndNotice{}  % leave blank if no need to mention equal contribution
\printAffiliationsAndNotice{\icmlEqualContribution} % otherwise use the standard text.

\begin{abstract}

We present the first unsupervised deep learning method for pollen analysis using bright-field microscopy. Using a modest dataset of 650 images of pollen grains collected from honey, we achieve family level identification of pollen. We embed images of pollen grains into a low-dimensional latent space and compare Euclidean and Riemannian metrics on these spaces for clustering. We propose this system for automated analysis of pollen and other microscopic biological structures which have only small or unlabelled datasets available.

\end{abstract}

\section{Introduction}
\label{submission}

Pollen is formed as the male gametes of all flowering plants. Palynology, the study of pollen (and spores), provides critical insights for a number of fields including forensic science, ecology and agriculture \cite{mildenhall_2006, blackmore_2006, zábrodská_vorlová_2014}. For example, pollen grains on the personal affects of an individual can be used to ascertain whether they were at the scene of a crime \cite{mildenhall_2006}. However, traditional methods of pollen analysis such as microscopic analysis of morphology are time intensive and require trained specialists. As a result, pollen analysis remains mostly inaccessible for the majority of large scale applications. Making the analysis of pollen fast, scalable and accessible can therefore open the door to a great many opportunities hitherto unfeasible in both commercial and academic domains. 

While automated methods of pollen analysis have been described in the literature as early as 2002 \cite{ronneberger_schultz_burkhardt_2002} the methods have not been implemented as significant tools in research or industry. Early attempts were largely scanning electron microscopy \cite{ronneberger_schultz_burkhardt_2002, treloar_taylor_flenley_2004} while later research used a wide variety of microscopy types including dark field \cite{lagerstrom_holt_arzhaeva_bischof_haberle_hopf_lovell_2014, pedersen_bailey_hodgson_holt_marsland_2017}, and fluorescence microscopy \cite{mitsumoto_yabusaki_aoyagi_2009}. All of these techniques require expensive equipment and most require relatively skilled operators. 

The introduction of deep learning techniques for pollen analysis has shown much promise, however, have been plagued by a lack of a significant labelled dataset. In a recent publication, a supervised deep learning algorithm was used to segment and then analyse pollen in honey samples \cite{he_gkantiragas_glowacki_2018}. However, the scope of this paper was limited by its requirement for the manual labelling of datasets. Here we provide the first example of the use of an unsupervised deep learning algorithm for the analysis of pollen using bright-field microscopy. 

\section{Methods}

\begin{figure}[ht]
\vskip 0.2in
\begin{center}
\centerline{\includegraphics[width=\columnwidth]{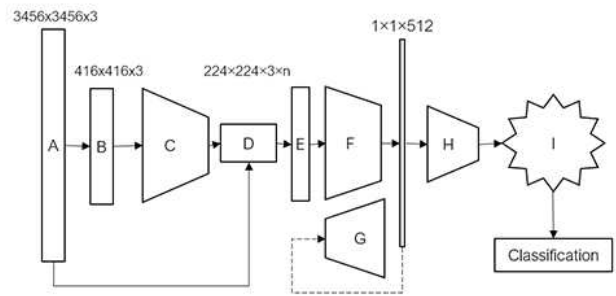}}
\caption{Overview of pollen identification. (A) Image of slide. (B) Downsampled image. (C) Object detection network. (D) Bounding boxes. (E) Full resolution pollen grain crops. (F) Encoder. (G) Decoder. (H) Further dimensionality reduction. (I) Cluster assignment.}
\label{arch}
\end{center}
\vskip -0.2in
\end{figure}

Unlike previous almost-entirely supervised approaches to pollen identification (which require large labelled datasets of pollen to train), we propose a new unsupervised pipeline (see Figure \ref{arch}) for generating representations of pollen and learning groupings of morphologically (and thererby often taxonomically \cite{oswald_doughty_neeman_neeman_ellison_2011}) similar pollen grains.

Given an input image of a microscope slide containing pollen, \(X \in \mathbb{R}_{+}^{3328 \times 3328 \times 3}\), we first apply downsampling such that \(X \in \mathbb{R}_{+}^{416 \times 416 \times 3}\). The downsampled \(X\) is then passed through a YOLO-based \cite{Redmon_2016_CVPR} object detection network from \cite{he_gkantiragas_glowacki_2018} to obtain a set of bounding boxes \(B \subset \mathbb{R}_{+}^{4} \times \mathbb{N}\) for pollen grains present on the slide containing indexed tuples \((x_{i}, y_{i}, w_{i}, h_{i})\) giving the positions and dimensions of the \(i\)th particular bounding box respectively. Each tuple in \(B\) is then scaled up component-wise by a factor of 8 and recombined with the original image to obtain centered full-resolution crops \(c_{i}\) of the pollen. Each \(c_{i}\) passed through an encoder based on the VGG16 architecture from \cite{vgg16} pre-trained on ImageNet \cite{deng_dong_socher_li_li_fei-fei_2009} with the final fully-connected layers replaced with a \(7 \times 7\) max-pooling layer resulting in a latent representation \(z_{i} \in Z \subset \R^{512}\). We then further reduce the dimensionality of \(z_{i}\) to  \(d_{final} \in \mathbb{N}\) and cluster the results using the k-means algorithm. In our experiments, we compare principal component analysis to the Isomap algorithm as dimensionality reduction methods as well as a Euclidean metric on \(Z\) to a Riemannian metric induced by the latent space. The Riemannian metric is approximated using a fixed-point algorithm described in \cite{Yang2018GeodesicCI} and reflects the curvature that arises from the non-linearity of the latent space which can be characterised as a Riemannian manifold  \cite{Arvanitidis2018LatentSO}. 

In this way, we are able to generate representations of the pollen without requiring a large dataset of pollen to learn from. Furthermore, the clusters discovered can now be used for a number of new tasks including semi-supervised learning for pollen classification.

\section{Experiments}
Samples from a range of honeys (eucalpytus melliodora, acacia and manuka) were spread onto a microscope slide and immediately imaged. Images were taken once every 2 seconds on a Solomark compound microscope at 320x zoom. This generated a dataset of 650 unlabelled bright-field images of individual pollen. 

\subsection{Clustering}
\begin{figure}[ht]
\vskip 0.2in
\begin{center}
\centerline{\includegraphics[width=\columnwidth]{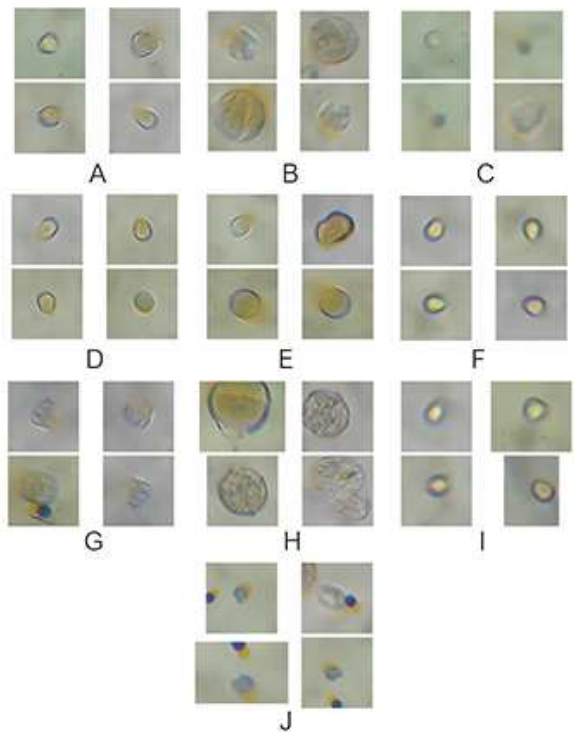}}
\caption{Randomly sampled members from different clusters of pollen imagery with PCA and a Euclidean metric on \(Z\).}
\label{pcaclusters}
\end{center}
\vskip -0.2in
\end{figure}

The system was run with parameters \(k = 10\) and  \(d_{final} = 3\) for ease of visualisation. The value of \(k\) was selected based on the ratio between decreases in cluster variance for adjacent values of \(k\). The system was in most cases able to differentiate pollen morphology on (at the minimum) a family level. For example, one cluster (see A in Figure \ref{pcaclusters}) shows a strong resemblance to pollen from the Myrtaceae family \cite{sniderman_matley_haberle_cantrill_2018}. 

\begin{figure}[ht]
\vskip 0.2in
\begin{center}
\centerline{\includegraphics[width=\columnwidth]{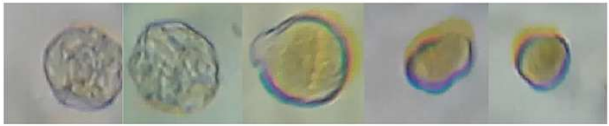}}
\caption{Samples along a principal component of \(Z\) seemingly encoding colour, surface texture and roundness.}
\label{pcaxis}
\end{center}
\vskip -0.2in
\end{figure}

\begin{figure}[ht]
\vskip 0.2in
\begin{center}
\centerline{\includegraphics[width=\columnwidth]{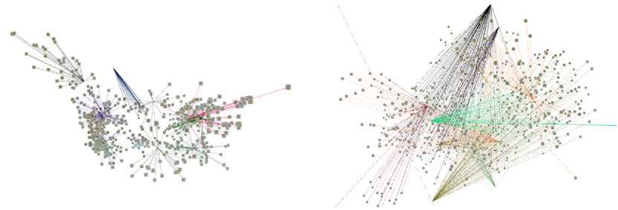}}
\caption{Latent space representations of Isomap (left) and PCA (right). Cluster assignments are illustrated with lines from each image point in the latent space to its respective centroid. Clusters are computed using a Euclidean metric on \(Z\) in both visualisations. Point size corresponds to depth.}
\label{latentspaces}
\end{center}
\vskip -0.2in
\end{figure}

\begin{figure}[ht]
\vskip 0.2in
\begin{center}
\centerline{\includegraphics[width=\columnwidth]{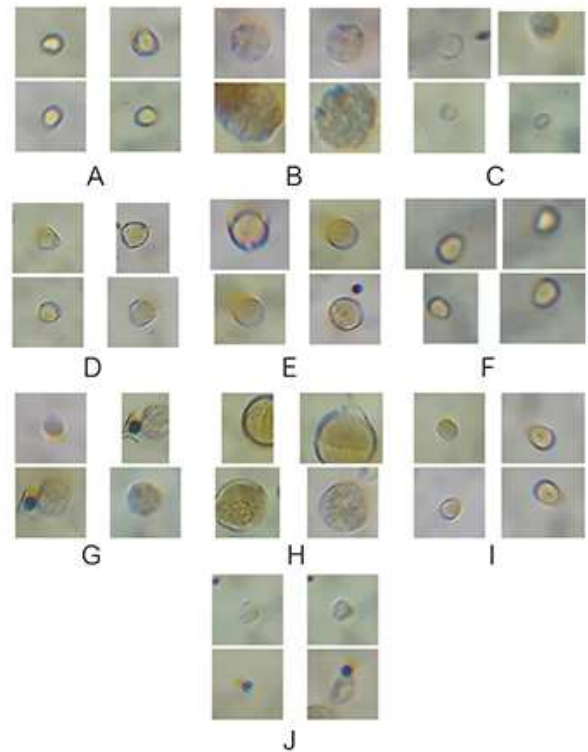}}
\caption{Randomly sampled members from different clusters of pollen imagery with Isomap and a Riemannian metric on \(Z\). }
\label{isoclusters}
\end{center}
\vskip -0.2in
\end{figure}

There are marked differences in the latent spaces generated by the two different dimensionality reduction methods. Qualitatively, clusters are far more clear in the spaces where Isomap was used (see Figure \ref{latentspaces}). Nonetheless, the embeddings in the PCA spaces were sensible (see Figure \ref{pcaxis}). The clusters were not too different in terms of overall concepts - the labels in Figure \ref{isoclusters} are matched to potential "conceptual counterparts" in Figure \ref{pcaclusters}. There were, however, qualitatively fewer obviously incorrect cluster assignments from the Isomap-Riemannian space compared to the other spaces.

\begin{figure}[ht]
\vskip 0.2in
\begin{center}
\centerline{\includegraphics[width=\columnwidth]{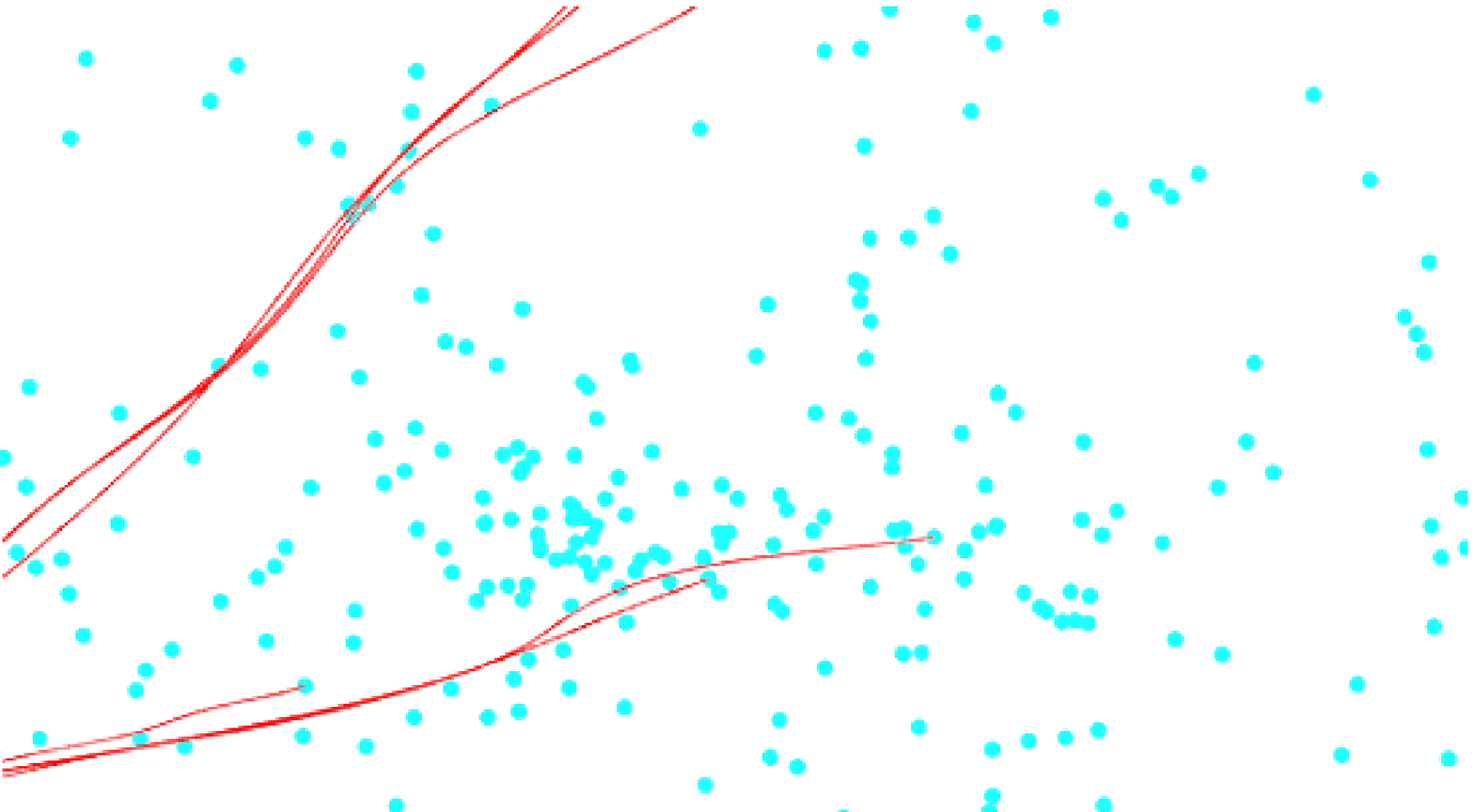}}
\caption{Visualising geodesics between random points on \(Z\) with Isomap and a Riemannian metric. The red lines are geodesics while the blue dots are image embeddings from the dataset. They are clearly non-linear over larger distances.}
\label{geodesics}
\end{center}
\vskip -0.2in
\end{figure}

We further visualise the curvature of the latent spaces by sampling geodesics between random points on each manifold. The geodesics observed on the PCA manifolds are (unsurprisingly) effectively linear while some more interesting curvature can be observed on the Isomap manifold in Figure \ref{geodesics}.

\subsection{Comparison with Human-Defined Cluster Assignments}

A set of 30 bright-field images were collected at 320x zoom using a combination of an AmScope bifocal microscope and a Solomark compound microscope. Non-specialist volunteers were instructed to categorise a random selection of these images into the already generated clusters given 4 randomly chosen images from each cluster (similar to Figure \ref{pcaclusters}). The images were then reclassified by the system (using PCA and an Euclidean metric on \(Z\)). Cases of agreement between the human and system was 63\%. Calculating Cohen's Kappa coefficient, we find moderate agreement \(\kappa = 0.576\) (95\% CI, 0.378 to 0.775).

\section{Discussion}

While previous research had used unsupervised methods to analyse pollen samples in grass \cite{mander_li_mio_fowlkes_punyasena_2013} we used a sample size which was significantly larger and did so with less expensive tools, making our method more accessible. Additionally, we were able to utilize a significantly larger dataset and observe an entire pollen grain, as opposed to simply surface patterns. Moreover, agreement between humans and our system was relatively high at 63\%. This implies that the clusters and assignments at large are to some extent human-interperatable. An important future benchmark would be comparison to human specialists who, with the help of reference material, are able to visually differentiate pollen at a species level with reasonable accuracy though such resolution is not necessary for many applications \cite{mildenhall_2006}.

While our system provides a new tool for working with small or unlabelled datasets, it has significant limitations. Primarily, the small sample size and poor quality of some images in our dataset limits the classification of pollen to the family level. A larger dataset and better equipment may be required to achieve genus and species level classification. Luckily, this should be relatively straightforward as no labels are needed. Alternatively, applying the same system used here on datasets generated using other microscopy types such as dark field or phase contrast microscopy could provide powerful new tools for rapid pollen analysis, though this would potentially not confer the same benefit with regards to accessibility. 

Although the latent spaces observed in the experiments were relatively sensible, often with the obvious clusters visible, it is likely that the embeddings are sub-optimal due to the pre-training of the encoder on ImageNet. This is due to the fact that biological imagery at the micro-scale is visually very different to much of the contents of ImageNet. It would therefore perhaps be beneficial to representation to use an encoder trained instead on a large dataset of microscope imagery. 

\subsection{Applications}
Beyond accelerating existing uses of pollen analysis, two socially beneficial applications of large-scale automated pollen analysis include combatting the proliferation of fraudulent honey and as well as ecological monitoring.

Fraudulent honey is a global issue and has seen honey become the third most faked food in worldwide \cite{moore_spink_lipp_2012}. Honey fraud is most commonly carried out through mislabelling, dilution with cheaper honeys or sugar syrup and a host of unethical beekeeping practices such as over-harvesting. Existing authentication methods for honey have been inaccurate (such as sugar or chemical testing) or highly specialist and expensive (such as qPCR, LC-MS or manual inspection of the pollen in honey \cite{mcdonald_keeling_brewer_hathaway_2018, doi:10.1021/jf501475h, sniderman_matley_haberle_cantrill_2018}). For this use case, we describe a system whereby producers of honey may upload scans of their honey taken through low-cost bright-field microscopy apparatus to an online database. With such a database, consumers and retailers down the supply chain will be able to verify the origin or authenticity of their honey from its pollen profile without specialist knowledge. This will help combat the harmful effects of fake honey on both the ecosystem and local economies described elsewhere \cite{fairchild, cairns_villanueva-gutierrez_koptur_bray_2005}. Furthermore, the system will have made up the cost of a microscope purchase within 3-4 jars of genuine manuka honey.

\begin{figure}[ht]
\vskip 0.2in
\begin{center}
\centerline{\includegraphics[width=\columnwidth]{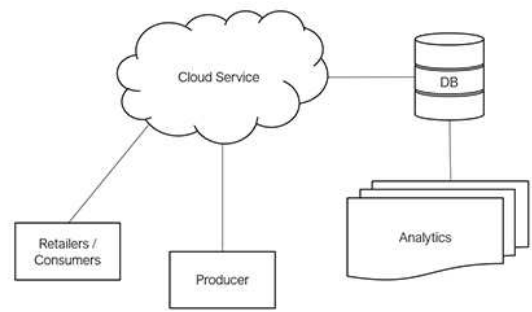}}
\caption{A potential architecture for ecosystem monitoring through large-scale honey authentication infrastructure. The database can not only be used for honey authentication, but also for long-term analytics on changing pollen distributions.}
\label{systemarch}
\end{center}
\vskip -0.2in
\end{figure}

In addition, we propose honey as a means for the large-scale monitoring of flora biodiversity. Previous studies have used honey as a way of monitoring pollutants \cite{smith_weis_amini_shiel_lai_gordon_2019} and conducting analysis on pollen would allow for simultaneous generation of data on flora biodiversity. This would go hand-in-hand with the honey authentication system described above, using the pollen profile database as a source of spatio-temporally tagged samples. Two caveats, however, ought not be overlooked: firstly, that this application would require a small quantity of expertly labelled pollen data in order to associate pollen grains with their respective botanical sources in a semi-supervised fashion; and secondly, that honey bees have specific foraging habits which would need to be corrected for. Nevertheless, pollen obtained from other sources such as dedicated collection units can be equally analysed by the system. We present a potential architecture for how both proposed applications can co-exist in Figure \ref{systemarch}. 

Finally, the system could be used for the analysis of any unlabelled microscopy dataset in the environmental and life sciences. Soil fungi, for example, are considered to be an excellent indicator of soil fertility \cite{kranabetter_friesen_gamiet_kroeger_2009}. The use of the system for such applications could accelerate the development of prototypes and proofs-of-concept by removing the need for a large labelled dataset.

\section{Conclusions}
Despite its limitations, the system we have described forms the groundwork for a powerful, scalable and accessible tool for pollen analysis. The acquisition of a larger dataset and an application-specific encoder are needed to bring the system to a level of maturity which, once reached, will open doors to novel techniques for ecological monitoring, honey authentication and a variety of other use cases. 

\section{Acknowledgements}
Many thanks to Stewart McGown at the University of St Andrews for assisting in the design and prototyping of the large-scale honey authentication system. We also acknowledge Raghvi Arya for her advice regarding the poster presentation at ICML. 

\bibliography{example_paper}
\bibliographystyle{icml2019}

\end{document}